\documentclass[conference]{IEEEtran}
\usepackage{times}
\usepackage[numbers,sort&compress]{natbib}
\usepackage{booktabs}
\usepackage{graphicx}
\usepackage{array}
\usepackage[bookmarks=true]{hyperref}
\usepackage{color} % Basic functionality
\usepackage{tabularx}
\usepackage{booktabs}
\newcolumntype{Y}{>{\raggedright\arraybackslash}X}
\newcommand{\worldobject}{\texttt{WorldObject}}
\newcommand{\skillplan}{\texttt{SkillPlan}}
\newcommand{\skillcall}{\texttt{SkillCall}}
\newcommand{\motiongoal}{\texttt{MotionGoal}}
\newcommand{\runtimefeedback}{\texttt{RuntimeFeedback}}

\begin{document}

\title{Representation Handoffs for OpenArm-Based Laboratory Mobile Manipulation}

\author{
\IEEEauthorblockN{
Yang Shen\IEEEauthorrefmark{1},
%Yang Shen\IEEEauthorrefmark{1}\IEEEauthorrefmark{2},
Chonghao Cheng\IEEEauthorrefmark{1},
Ziyi Zhao\IEEEauthorrefmark{1},
Jialuo Zhu\IEEEauthorrefmark{2},
Zhenyi Yi\IEEEauthorrefmark{2},\\
Qi Zhao\IEEEauthorrefmark{2},
Jian Yang\IEEEauthorrefmark{2},
Yuhui Shi\IEEEauthorrefmark{2},
Chin-Teng Lin\IEEEauthorrefmark{1}
}
\IEEEauthorblockA{
\IEEEauthorrefmark{1}University of Technology Sydney\\
%Email: chin-teng.lin@uts.edu.au
}
\IEEEauthorblockA{
\IEEEauthorrefmark{2}Southern University of Science and Technology\\
%Email: shiyh@sustech.edu.cn
}
}

\maketitle

\begin{abstract}
% Open-source robot arms and foundation-model perception have lowered the barrier to building embodied intelligence prototypes, but deploying them for language-based laboratory manipulation still requires a reliable path from instructions and sensor observations to safe robot actions.
Open-source robotics and foundation models have lowered the barrier to embodied AI, yet language-guided laboratory automation still requires reliable alignment from instructions and observations to safe actions. This field report presents an OpenArm-based mobile manipulation prototype for laboratory-style tasks, built by integrating dual OpenArm manipulators with a mobile base, vertical slide, RGB-D sensing, lidar-based mapping, ROS2/MoveIt execution, and profile-defined skill interfaces. The system is organized around representation handoffs: natural language requests are constrained into registered skill calls, sensor observations are grounded into maps and object poses, object priors provide role and skill constraints, and runtime bindings compile validated skills into executable motion goals. We use dry-run traces and startup checks to evaluate this integration path, showing how the prototype exposes missing calibration, incomplete object assets, and unfinished real scene visual grounding as explicit deployment blockers. These intermediate representations serve as practical debugging interfaces for integrating language, perception, planning, and robot safety in embodied systems. Code is available at \url{https://github.com/yshenfab/xEI}.
\end{abstract}

\IEEEpeerreviewmaketitle

\section{Introduction}
\begin{figure}[!htbp]
\centering
\includegraphics[width=0.45\textwidth]{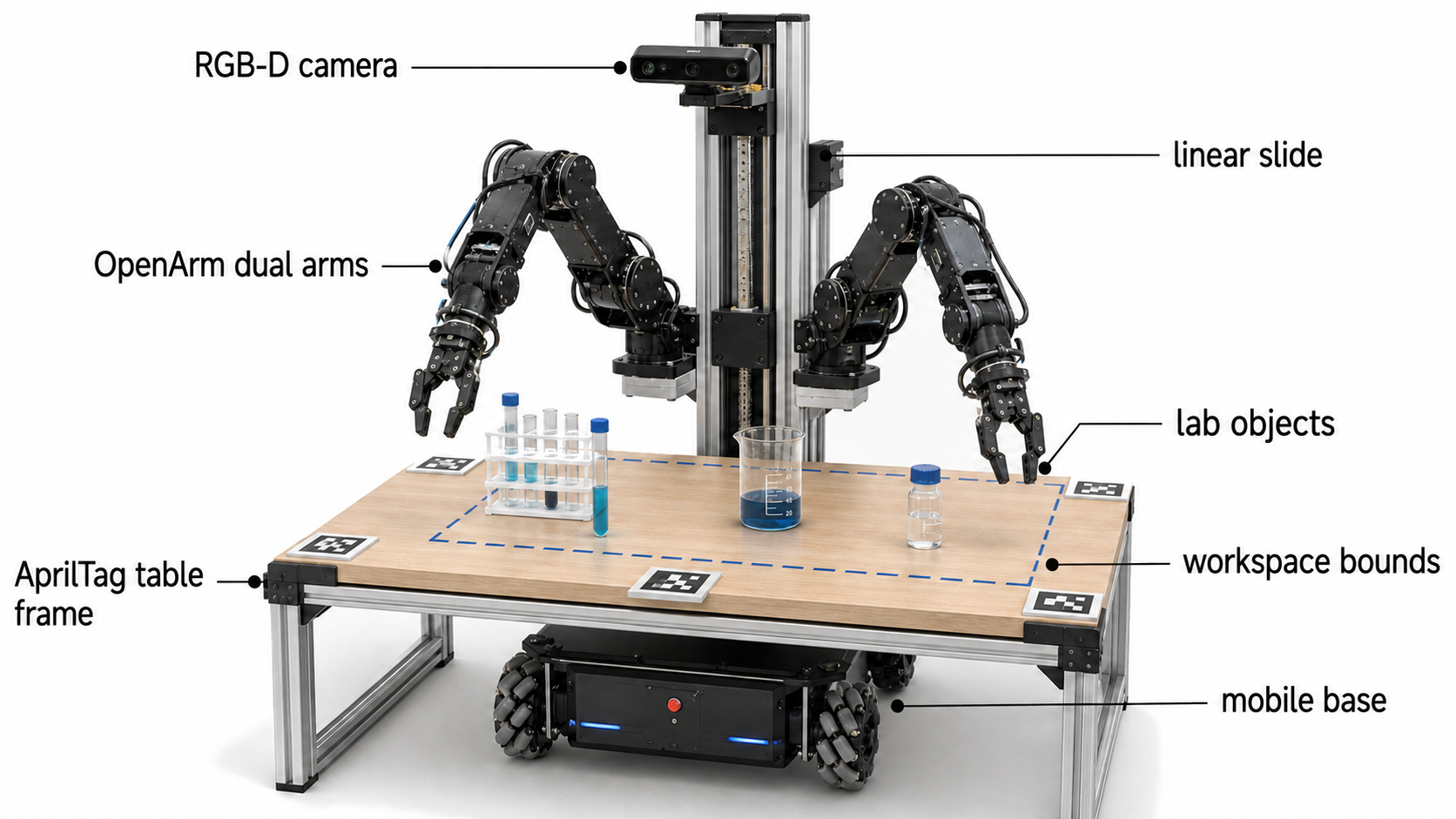}
\caption{Our OpenArm-based prototype for language-based laboratory manipulation. Our system targets tasks in which natural-language instructions must be grounded to laboratory objects and executed as manipulation skills, including moving containers, picking and placing items, handling tubes or racks, and preparing actions such as pouring or sample transfer.}
\label{fig:system}
\end{figure}

Open-source robot hardware, ROS-based middleware, and foundation model perception have made it easier to assemble embodied intelligence prototypes, but turning such prototypes into reliable task-execution systems remains difficult. In this field report, we introduce an OpenArm-based mobile manipulation prototype for laboratory-style embodied tasks. Our mobile manipulation prototype is built on OpenArm\footnote{\url{https://openarm.dev/}}, a low-cost, open-source humanoid arm platform. The prototype integrates dual OpenArm manipulators, grippers, a vertical slide, a mobile base, RGB-D sensing, lidar-based mapping, ROS2/MoveIt execution, and profile-defined skill interfaces for laboratory-style manipulation tasks. The prototype is shown in Fig.~\ref{fig:system}. For mobile manipulation under language-based lab instructions, the main integration question is therefore not only how to represent the scene, but which representation is actionable by a real robot.

The workflow can be described as follows. A user gives a natural-language laboratory task, the robot moves to an appropriate scene origin, grounds relevant objects, asks a large language model (LLM) planner to emit for registered skill calls, validates the result against a profile-defined skill bank, expands each call into robot frame motion goals, and records the resulting trace. This workflow cannot be implemented safely by passing an LLM directly to robot control, and a 6D object pose is not enough either: it must be paired with object roles, skill contracts, frames, geometric operation parameters, safety limits, and execution feedback. We implement the OpenArm-based mobile manipulation prototype through representation handoffs: laboratory instructions are represented by validated skill-call sequences; lidar and RGB-D observations are represented by maps and object poses; object priors are represented by skill arguments; and runtime bindings are represented by executable motion goals.

\begin{figure*}[t]
\centering
\begin{minipage}[t]{0.19\textwidth}
\centering
\includegraphics[width=\linewidth,height=0.24\textheight,keepaspectratio]{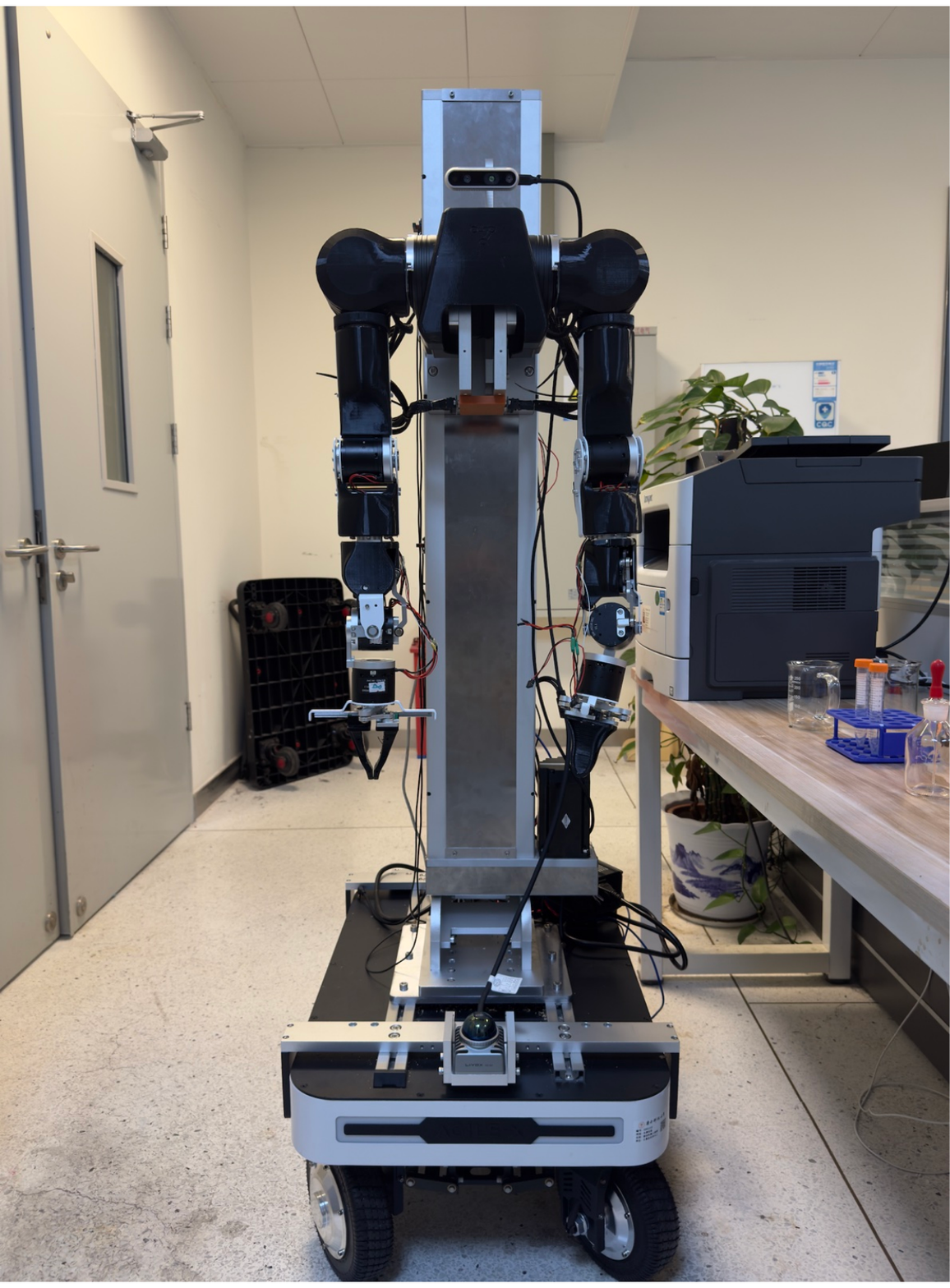}\\[-1mm]
\footnotesize (a) Prototype body
\end{minipage}\hfill
\begin{minipage}[t]{0.19\textwidth}
\centering
\includegraphics[width=\linewidth,height=0.24\textheight,keepaspectratio]{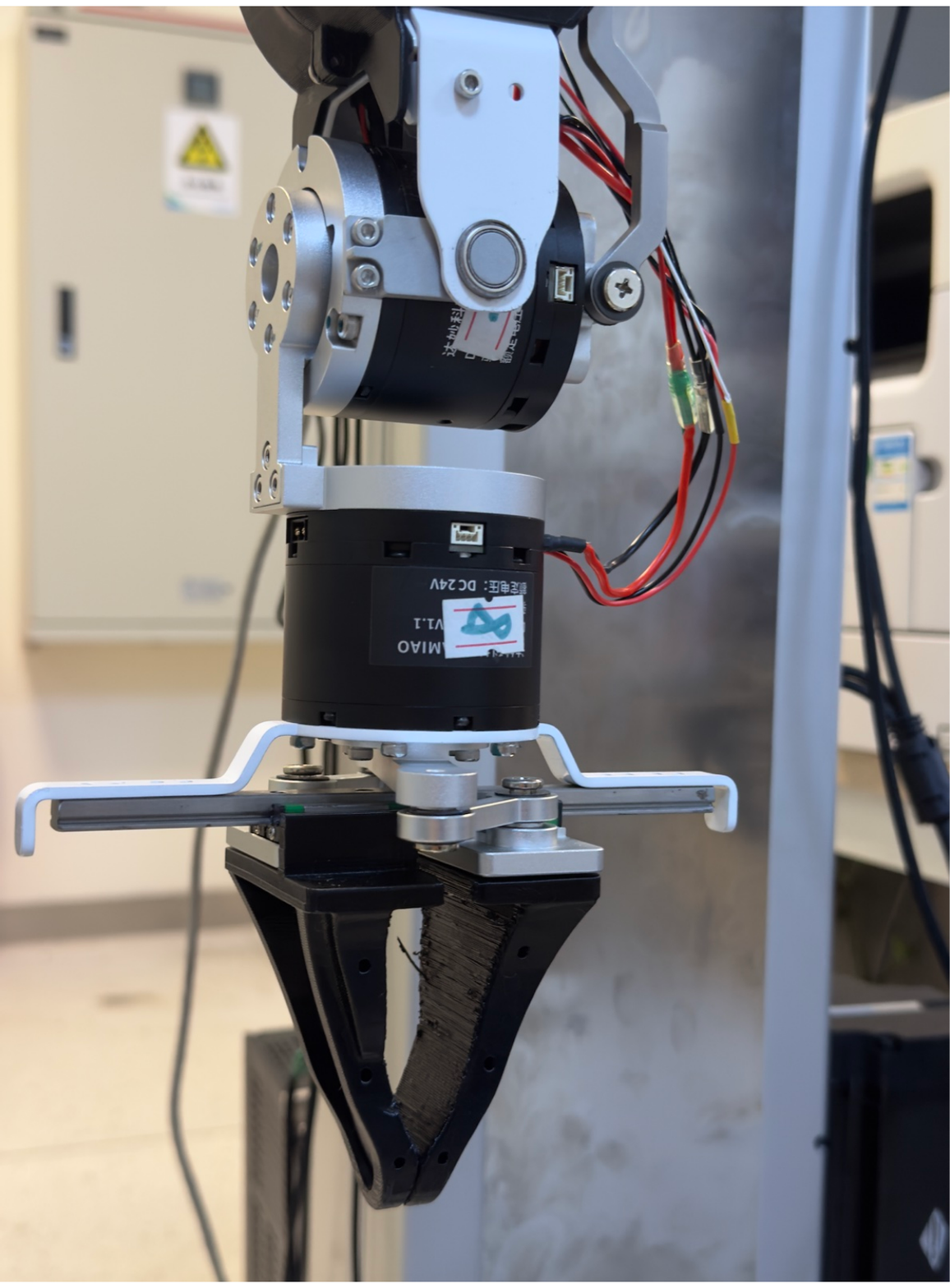}\\[-1mm]
\footnotesize (b) Arm gripper
\end{minipage}\hfill
\begin{minipage}[t]{0.19\textwidth}
\centering
\includegraphics[width=\linewidth,height=0.24\textheight,keepaspectratio]{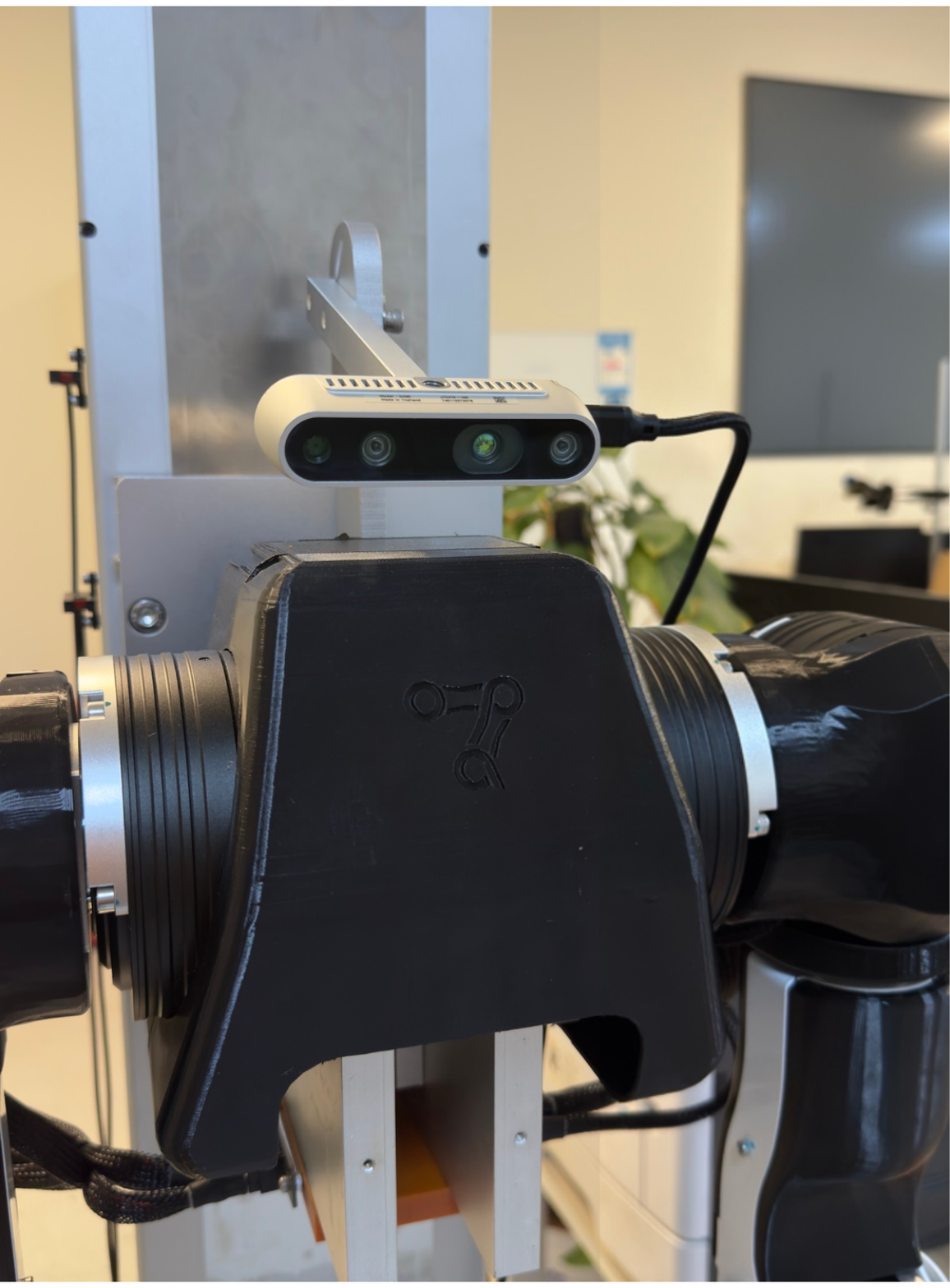}\\[-1mm]
\footnotesize (c) RGB-D camera
\end{minipage}\hfill
\begin{minipage}[t]{0.19\textwidth}
\centering
\includegraphics[width=\linewidth,height=0.24\textheight,keepaspectratio]{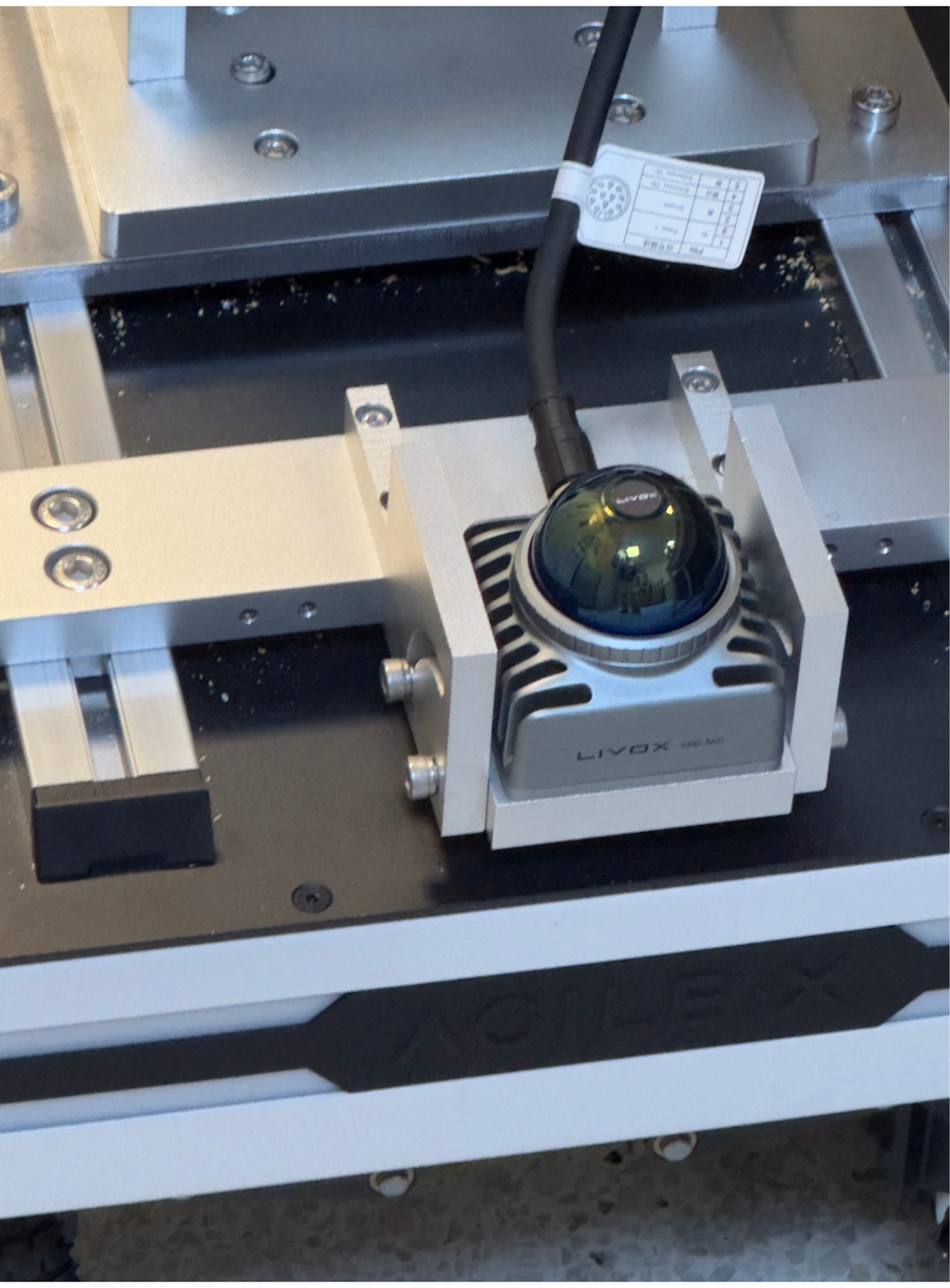}\\[-1mm]
\footnotesize (d) Lidar
\end{minipage}
\begin{minipage}[t]{0.19\textwidth}
\centering
\includegraphics[width=\linewidth,height=0.24\textheight,keepaspectratio]{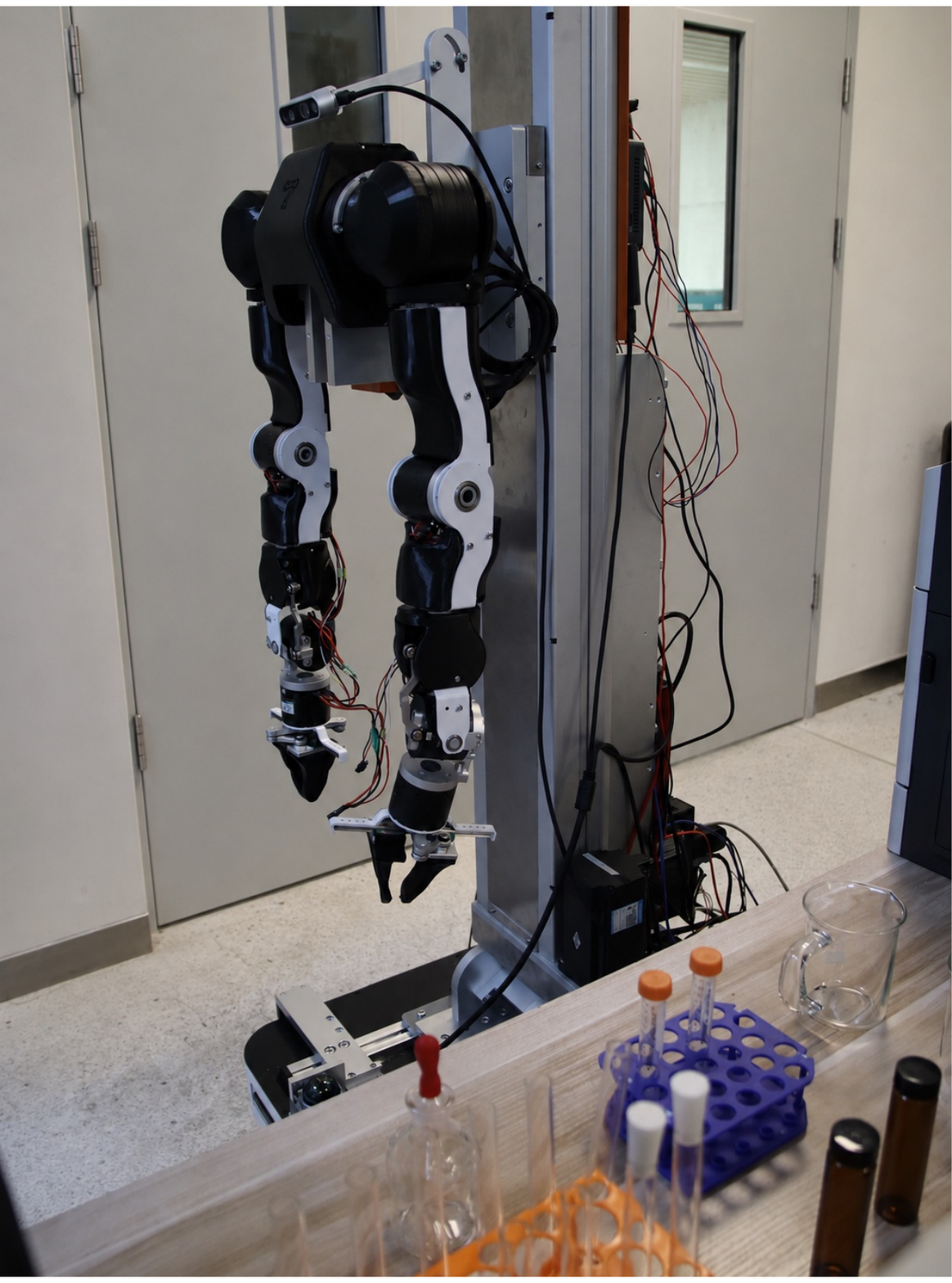}\\[-1mm]
\footnotesize (e) Scene
\end{minipage}
\caption{Hardware components of the OpenArm-based mobile manipulation prototype. (a) The prototype body integrates the OpenArm dual arms, vertical slide, and mobile base. (b) The arm gripper provides the end-effector interface for manipulation skills. (c) The RGB-D camera provides tabletop observations for object grounding and pose estimation. (d) The lidar supports mobile base mapping and navigation. (e) The application scenario.}
\label{fig:hardware-components}
\end{figure*}

The central observation from our deployment is that these interfaces need to be treated as explicit representations. We therefore frame the prototype around representation handoffs: instructions are constrained into registered skill calls, sensing outputs are grounded into frames and object states, object priors specify roles and admissible actions, and validated skills are bound to executable motion goals. This field report frames the prototype as a representation-centric system integration effort. The main contributions are summarized as follows:
\begin{enumerate}
\item A system view of the representation handoffs among lab instructions, maps, object poses, object priors, registered skill calls, runtime bindings, and motion goals;
\item An OpenArm-based integration that connects ROS2/MoveIt execution, mobile base navigation, vertical motion, RGB-D grounding, a skill bank, and deterministic skill templates;
\item Dry-run evidence and field lessons showing where missing calibration, object models, perception outputs, and capability contracts block real-robot deployment.
\end{enumerate}
This system field report includes dry runs and mock perception traces, startup checks, and integration lessons.

\section{System Overview}

We built an OpenArm-based mobile manipulation prototype for laboratory-style embodied tasks. The main hardware components are shown in Fig.~\ref{fig:hardware-components}. The current software stack uses ROS2 middleware~\cite{quigley2009ros,macenski2022robot}, MoveIt/OpenArm arm execution~\cite{chitta2012moveit,sucan2012open}, linear slide, mobile base navigation, FoundationPose model~\cite{wen2024foundationpose}, and AprilTag~\cite{olson2011apriltag}. In our embodied system, the top-level agent is a deterministic orchestrator; the LLM is used only by a unified planner that must emit registered \skillcall{} records against the development profile skill bank. If the planner fails validation or cannot cover a requested capability, the system reports \texttt{unmapped\_requests}. Physical execution is routed through runtime bindings, safety checks, and ROS2/OpenArm backends.

\begin{figure*}[t]
\centering
\includegraphics[width=0.95\textwidth]{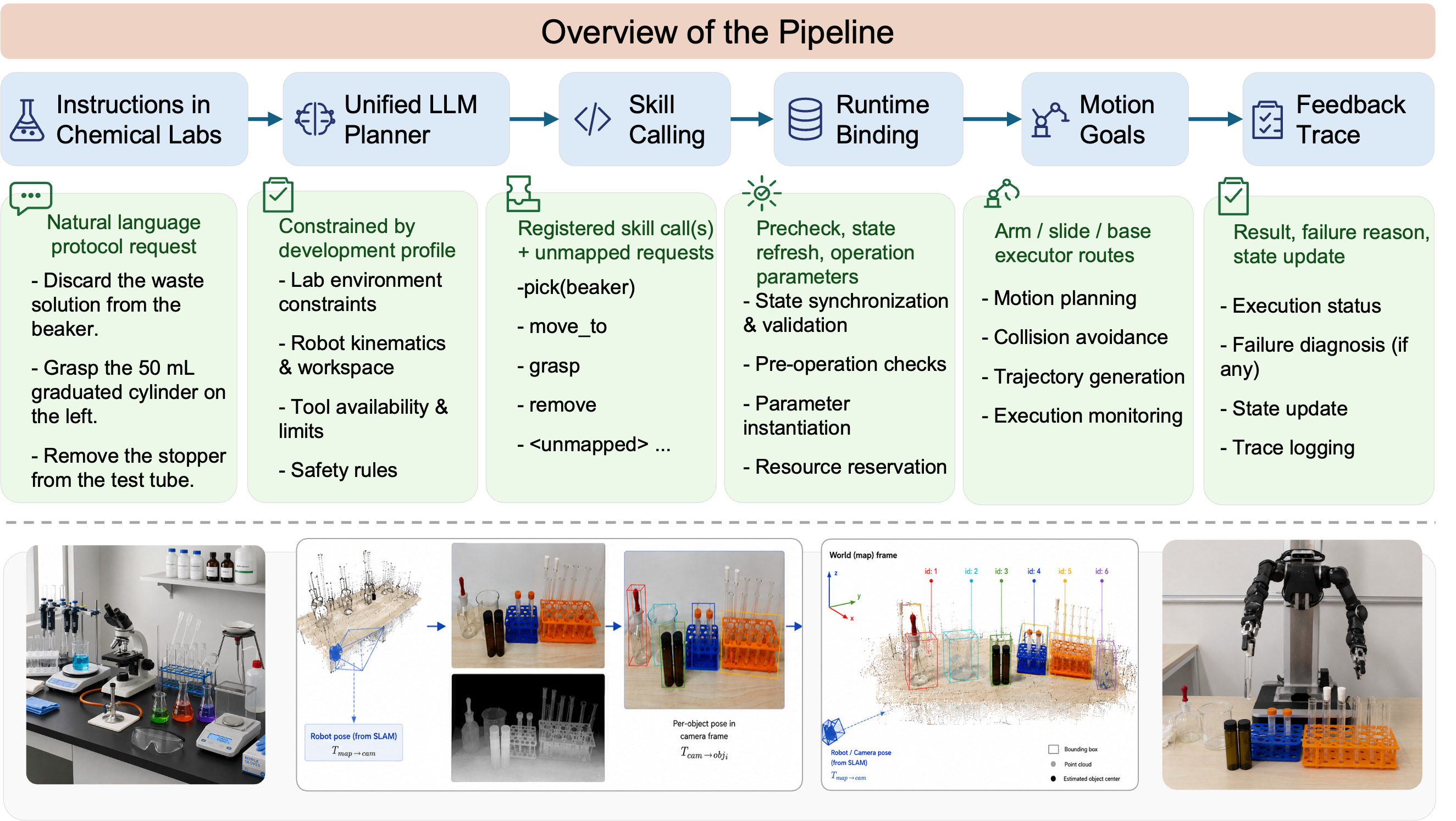}
\caption{Overview of the proposed language grounded laboratory manipulation pipeline. A unified LLM planner converts natural language instructions into executable robot skills through runtime binding and motion planning, while feedback traces monitor execution and update system states. Representative examples of world grounding and robotic operation are shown at the bottom.}
\label{fig:handoffs}
\end{figure*}

Figure~\ref{fig:handoffs} summarizes the pipeline and data flow. Before grounding the world, the robot may move to a configured scene origin. World grounding then combines usage-profile object priors, system objects, perception, or a FoundationPose response. The resulting \worldobject{} list, current robot state, planner-facing skill bank, and memory context form the input to a unified planner. The planner returns a \skillplan{} containing ordered \skillcall{} records, explicit \texttt{unmapped\_requests}, and safety notes. The output is validated against the skill bank, object role requirements, argument types, and step-id format; one validation-repair attempt is allowed before the request is reported as unmapped. Finally, the embodied runtime pre-checks the plan, refreshes the robot and world states before each skill call, resolves object- and target-specific operation parameters, validates safety limits, and executes a sequence of \motiongoal{} records. For more details, please refer to section~\ref{sec:rep} and our code repository.

This structure sits between classic robot middleware and recent language-based robotics.
LLM and Vision-Language-Action (VLA) systems demonstrate powerful task-level priors and generalization~\cite{ahn2022can,driess2023palm,zitkovich2023rt,liang2023code,huang2023voxposer}.
% Our prototype takes a conservative integration stance: language can choose among registered capabilities, but it does not directly bypass the contracts that turn instructions into robot actions.
%Figure~\ref{fig:system} shows the annotated system view behind this abstraction, while Fig.~\ref{fig:hardware-components} shows the physical actuation and sensing components that support the representation handoffs.

\section{Task-aware Representation Pipeline}
\label{sec:rep}
\subsection{Environment Representation}
%Spatial Frames, Maps, and Table Anchors
In our laboratory manipulation setting, the first representation problem is spatial: before the robot can ground objects or execute a skill, it must know how the navigation map, tabletop workspace, RGB-D camera, and OpenArm base relate to one another. Laser scans are accumulated into a navigation map for mobile base motion~\cite{thrun2002probabilistic,cadena2016past}, while RGB-D observations are expressed in calibrated camera and robot frames for manipulation. The calibration profile defines the coordinate frames used by the system, including the world, robot base, MoveIt base, tool, camera, lidar, slide, and table frames. It also records the static transforms and verification tolerances used to check frame alignment. Sensor topics, table bounds, and workspace bounds are treated as deployment constraints rather than implicit assumptions. AprilTag detections are used as table anchors: given detected tag poses and a known tag-to-table layout, the system can estimate or verify the tabletop frame consumed by downstream grounding and manipulation modules. Hand/eye and frame calibration are well-studied problems~\cite{tsai1989new}, but in a field system, they become operational representations. This spatial handoff makes deployment assumptions explicit, so that a dry-run or startup trace can explain which map frame, table frame, camera transform, and workspace bounds were used before object grounding or motion execution began.

\subsection{Object Roles and Skill Interfaces}
%Object Priors as Skill Interfaces
The second handoff is from laboratory objects to skill-facing symbols.
The current object registry separates system objects from day-to-day laboratory objects: the registry keeps system boundaries and a usage-profile pointer, while \texttt{usage\_profile.yaml} stores the current laboratory object priors. Each laboratory object is represented by a name, a point cloud or mesh path, aliases, physical dimensions, semantic roles, and exposed skills. For example, a tube can be represented as movable, a liquid container, a reaction vessel, an insertion target, and a sample container; a rack can be a fixture, a surface, and a place target. This is related to the affordance view of perception, where the meaning of an object is grounded in possible actions~\cite{gibson1979ecological,bohg2013data}. For our prototype, affordances are deliberately engineered as explicit roles and skill memberships before they are learned. The development profile then exposes a skill bank: each skill declares an owner, target role requirements, typed arguments, preconditions, unsupported requests, and internal process notes.

\subsection{Object Poses and Operation Parameters}

The third handoff converts perception into runtime-facing objects.
A \worldobject{} contains a 6D pose, frame id, confidence, roles, skills, and metadata.
The FoundationPose service boundary is now a concrete HTTP/RGB-D protocol: the agent sends target queries, an output frame, camera intrinsics, RGB, depth, and optional masks; the GPU-side service returns object names, frames, confidence, skills, roles, poses, and mesh metadata~\cite{wen2024foundationpose}.
The service also exposes health, model, and object endpoints, with a mock mode for software integration before the GPU perception stack and masks are available.

The prototype then enriches \worldobject{} records with deterministic operation parameters.
Given pose, computed size, roles, available skills, and runtime bindings, the operation-parameter planner generates approach, grasp, lift, place, insert, pour, dispense, tap, swipe, and twist parameters.
These defaults are stored as \texttt{perception\_args}, \texttt{skill\_args}, and \texttt{target\_skill\_args}, then merged by the runtime when compiling a skill call.
This handoff serves as an interpretable bridge between perception outputs and skill templates, so that a dry-run trace can explain exactly why a particular pick pose, place pose, insertion depth, or pour clearance was chosen.

\subsection{Goal and Task Representation}
%Instructions, Skills, and Motion Goals

The final handoffs are from language to registered skills and from skill structure to motion. A \skillplan{} stores registered \skillcall{} records, explicit \texttt{unmapped\_requests}, and safety notes. Validation checks the selected skill, owner, target role requirements, typed arguments, object references, and step-id ordering. The runtime then applies skill-runtime bindings and expands each skill call into one or more \motiongoal{} records such as \texttt{move\_to}, \texttt{open\_gripper}, \texttt{set\_do}, \texttt{slide\_move}, or \texttt{nav\_to\_pose}. Every motion goal passes through a safety gate and is then routed to the appropriate executor. Before execution and each skill call, the runtime checks required objects, low-confidence objects, held-object preconditions, current robot state, and refreshed world grounding. If an object is missing, below confidence, violates a held-object precondition, or fails during execution, the runtime records \runtimefeedback{} and stops.

This stop-on-failure policy is deliberately conservative. It prevents a language planner from papering over missing perception or unsupported skills. It also makes the representation handoffs easy to debug, i.e., a failure can be localized to scene origin preparation, world grounding, planner validation, skill coverage, operation-parameter resolution, runtime precheck, safety validation, executor routing, or hardware execution.

\section{Pipeline Execution}

Table~\ref{tab:trace} summarizes the current evidence level. In a dry run, the system can execute all representation handoffs and generate a complete executable trace. For a simple transfer instruction with a valid LLM plan, the planner emits a registered \texttt{move\_object} call against a movable source and a destination object satisfying target-role requirements, and the runtime expands it into a pick-place sequence. When the LLM is unavailable, or the request falls outside the skill bank, the same interface produces explicit \texttt{unmapped\_requests}. The trace contains not only the final motion goals but also the world snapshot, grounded object metadata, resolved operation parameters, robot state, runtime precheck feedback, and executor feedback.

\begin{table}[t]
\centering
\caption{Dry-run trace structure for the instruction
``put source object on destination object.''}
\label{tab:trace}
\footnotesize
\setlength{\tabcolsep}{3pt}
\renewcommand{\arraystretch}{1.08}

\begin{tabularx}{\columnwidth}{
@{}
>{\bfseries}l
>{\raggedright\arraybackslash}X
>{\raggedright\arraybackslash}X
@{}}
\toprule
\textbf{Stage} &
\textbf{Representation} &
\textbf{Observed dry-run} \\
\midrule

Scene setup &
scene origin \motiongoal{} &
Navigation target in map frame before grounding \\

World state &
\worldobject{} list &
Lab objects plus system objects with roles, skills, and frames \\

Planning &
\skillplan{} plus capability gaps &
Registered skill-call sequence, or explicit gap when LLM access or validation fails \\

Runtime \\precheck &
required objects, held-object state, safety gate &
Target/object availability and goal safety checked before execution \\

Runtime \\execution &
operation parameters and \motiongoal{} records &
pick-above, pick-down, gripper close, lift, place, release, retreat \\

Feedback &
\runtimefeedback{} &
Per-goal robot state, world snapshot, resolved args, executor result \\

\bottomrule
\end{tabularx}
\end{table}

Startup checks also expose the current deployment boundary.
The profile files are parseable, and the dry-run execution path is available, but the real-robot visual loop still depends on recording first-time calibration, replacing placeholder camera/lidar/slide transforms, populating object mesh and point-cloud assets, measuring table height, and enabling the FoundationPose service with real RGB-D input and masks or a detector. At this stage, the system validates the representation pipeline and software integration; currently, it does not provide real-world success rates for visual grasping, pouring, or insertion into evaluation.

\section{Field Lessons}

\textbf{Lesson 1: 6D pose is necessary but insufficient.}
The outputs of FoundationPose are valuable, but a mobile manipulation system needs pose plus frame, confidence, object identity, geometry, skills, roles, and target-specific operation parameters.
The downstream planner should see an actionable object, not a raw perception result.

\textbf{Lesson 2: Profiles are representations, not merely configurations.}
The usage, calibration, and development profiles encode three different forms of state: object priors, geometric deployment facts, and capability contracts.
Treating these profiles as first-class representations made startup checks and dry-run traces more informative than ad hoc configuration files.

\textbf{Lesson 3: Constrained intermediate forms make LLM planning debuggable.}
The LLM only proposes registered skill calls under a skill bank contract.
It cannot invent arbitrary ROS commands, bypass role requirements, use undeclared arguments, or execute unregistered recovery.
This design sacrifices flexibility, but it makes unsupported capabilities visible as \texttt{unmapped\_requests} rather than hidden model assumptions.

\textbf{Lesson 4: Deployment blockers are representation blockers.}
The remaining real-robot work is not just hardware bringup.
It is the completion of missing representations: measured camera/lidar/slide transforms; table frame localization; object meshes; real RGB-D masks or detections; and calibrated workspace bounds.
The field report value is that these blockers appear at the representation interfaces, where they can be tested.

\section{Conclusion}

This report reframes an OpenArm-based embodied intelligence prototype as a set of representation handoffs across system integration. The central artifact is not a new perception model or controller, but an auditable path from laboratory instructions, maps, object poses, object priors, and skill bank contracts to validated skill calls, operation parameters, motion goals, and runtime feedback. The current implementation provides software-level and dry-run evidence that these handoffs are executable, while also making clear what remains before real-scene visual manipulation can be claimed. The next step is to replace placeholder calibration and model assets with measured field data, align the FoundationPose service with the current usage-profile object registry, enable strict real-scene grounding, and evaluate task-level failures across pick, place, insert, pour, clean, and screen-interaction tasks.

\section*{Acknowledgments}

This work is supported by the Shenzhen Fundamental Research Program under Grant No. JCYJ20200109141235597, National Natural Science Foundation of China under Grants No. 72401122 and 61761136008, Guangdong Basic and Applied Basic Research Foundation under Grants No. 2024A1515012241 and 2021A1515110024, Shenzhen Peacock Plan under Grant No. KQTD2016112514355531, Program for Guangdong Introducing Innovative and Entrepreneurial Teams under Grant No. 2017ZT07X386, and the Australian Research Council (ARC) under Discovery Grant DP210101093 and DP220100803.

\bibliographystyle{plainnat}
\bibliography{references}

\end{document}